\documentclass[11pt]{article}
\usepackage{eacl2017}
\usepackage{times}
\usepackage{url}
\usepackage{latexsym}
\usepackage{multirow}
\usepackage[T1]{fontenc}
\usepackage{graphicx}
\usepackage{subfigure}
\usepackage{color}
\usepackage{amsfonts}
\usepackage{amsmath}
\usepackage{comment}
\usepackage{todonotes}
\usepackage{algorithm2e}

\eaclfinalcopy 
\def\eaclpaperid{375} 

\title{Tackling Error Propagation through Reinforcement Learning:\\A Case of Greedy Dependency Parsing}

\author{Minh L{\^e}\\CLTL\\
	    Vrije Universiteit Amsterdam\\
        Amsterdam, The Netherlands\\
	    {\tt m.n.le@vu.nl}
	  \And
	Antske Fokkens\\CLTL\\
	    Vrije Universiteit Amsterdam\\
        Amsterdam, The Netherlands\\
  {\tt antske.fokkens@vu.nl} }

\date{}

\begin{document}

\maketitle

\begin{abstract}
Error propagation is a common problem in NLP. Reinforcement learning explores erroneous states during training and can therefore be more robust when mistakes are made early in a process. In this paper, we apply reinforcement learning to greedy dependency parsing which is known to suffer from error propagation. Reinforcement learning improves accuracy of both labeled and unlabeled dependencies of the Stanford Neural Dependency Parser, a high performance greedy parser, while maintaining its efficiency. We investigate the portion of errors which are the result of error propagation and confirm that reinforcement learning reduces the occurrence of error propagation.
\end{abstract}

\section{Introduction}

Error propagation is a common problem for many NLP tasks \cite{song12,quirk.corstonoliver06,han2013effects,gildea.palmer02,yang.cardie13}. It can occur when NLP tools applied early on in a pipeline make mistakes that have negative impact on higher-level tasks further down the pipeline. It can also occur within the application of a specific task, when sequential decisions are taken and errors made early in the process affect decisions made later on.

When reinforcement learning is applied, a system actively tries out different sequences of actions. Most of these sequences will contain some errors. We hypothesize that a system trained in this manner will be more robust and less susceptible to error propagation.

We test our hypothesis by applying reinforcement learning to greedy transition-based parsers \cite{Yamada2003,W04-0308}, which have been popular because of superior efficiency and accuracy nearing state-of-the-art. They are also known to suffer from error propagation. Because they work by carrying out a sequence of actions without reconsideration, an erroneous action can exert a negative effect on all subsequent decisions. By rendering correct parses unreachable or promoting incorrect features, the first error induces the second error and so on. \newcite{mcdonald.nirve07} argue that the observed negative correlation between parsing accuracy and sentence length indicates error propagation is at work.


We compare reinforcement learning to supervised learning on \newcite{chen.manning.2014}'s parser. This high performance parser is available as open source. It does not make use of alternative strategies for tackling error propagation and thus provides a clean experimental setup to test our hypothesis. Reinforcement learning increased both unlabeled and labeled accuracy on the Penn TreeBank and German part of SPMRL \cite{Seddah2014IntroducingLanguages}. This outcome shows that reinforcement learning has a positive effect, but does not yet prove that this is indeed the result of reduced error propagation. We therefore designed an experiment which identified which errors are the result of error propagation. We found that around 50\% of avoided errors were cases of error propagation in our best arc-standard system. Considering that 27\% of the original errors were caused by error propagation, this result confirms our hypothesis. 

This paper provides the following contributions:
\begin{enumerate}
\item We introduce Approximate Policy Gradient (APG), a new algorithm that is suited for dependency parsing and other structured prediction problems.
\item We show that this algorithm improves the accuracy of a high-performance greedy parser.
\item We design an experiment for analyzing error propagation in parsing.
\item We confirm our hypothesis that reinforcement learning reduces error propagation.
\end{enumerate}

To our knowledge, this paper is the first to experimentally show that reinforcement learning can reduce error propagation in NLP. 

The rest of this paper is structured as follows. We discuss related work in Section~\ref{sec:relwork}. This is followed by a description of the parsers used in our experiments in Section~\ref{sec:ourparser}. Section~\ref{sec:reinforcement-experiments} outlines our experimental setup and presents our results. The error propagation experiment and its outcome are described in Section~\ref{sec:errorpropexp}. Finally, we conclude and discuss future research in Section~\ref{sec:conclusion}.

\section{Related Work}\label{sec:relwork}

In this section, we address related work on dependency parsing, including alternative approaches for reducing error propagation, and reinforcement learning.

\subsection{Dependency Parsing}

We use \newcite{chen.manning.2014}'s parser as a basis for our experiments. Their parser is open-source and has served as a reference point for many recent publications \cite[among others]{Dyer2015,weiss15,alberti-EtAl:2015:EMNLP,honnibal-johnson:2015:EMNLP}. 
They provide an efficient neural network that learns dense vector representations of words, PoS-tags and dependency labels. This small set of features makes their parser significantly more efficient than other
popular parsers, such as the Malt \cite{niv:hal:nil:07} or MST 
\cite{mcdon:per:rib:haj:05} parser
while obtaining higher accuracy. 
They acknowledge the error propagation problem of greedy parsers, but leave addressing this through (e.g.) beam search for future work.

\newcite{Dyer2015} introduce an approach that uses Long Short-Term Memory (LSTM). Their parser still works incrementally and the number of required operations grows linearly with the length of the sentence, but it uses the complete buffer, stack and history of parsing decisions, giving the model access to global information. 
\newcite{weiss15} introduce several improvements on \newcite{chen.manning.2014}'s parser. Most importantly, they put a globally-trained perceptron layer instead of a softmax output layer. Their model uses smaller embeddings, rectified linear instead of cubic activation function, and two hidden layers instead of one. They furthermore apply an averaged stochastic gradient descent (ASGD) learning scheme. In addition, they apply beam search and increase training data by using unlabeled data through the tri-training approach introduced by \newcite{li:zha:che:14}, which leads to further improvements.

\newcite{kiperwasser.goldberg16} introduce a new way to represent features using a bidirectional LSTM and improve the results of a greedy parser. \newcite{Andor2016} present a mathematical proof that globally normalized models are more expressive than locally normalized counterparts and propose to use global normalization with beam search at both training and testing.

Our approach differs from all of the work mentioned above, in that it manages to improve results of \newcite{chen.manning.2014} \textit{without changing the architecture of the model nor the input representation}. The only substantial difference lies in the way the model is trained. In this respect, our research is most similar to training approaches using dynamic oracles \cite{Goldberg2012}. Traditional static oracles can generate only one sequence of actions per sentence. A dynamic oracle gives \textit{all} trajectories leading to the best possible result from every valid parse configuration. They can therefore be used to generate more training sequences including those containing errors. A drawback of this approach is that dynamic oracles have to be developed specifically for individual transition systems (e.g.\ arc-standard, arc-eager). 
Therefore, a large number of dynamic oracles have been developed in recent years \cite{Goldberg2012,goldberg.nirve13,goldberg2014,gomez2014,Bjorkelund2015}.
In contrast, the reinforcement learning approach proposed in this paper is more general and can be applied to a variety of systems.

\newcite{zhang.chan.2009} present the only study we are aware of that also uses reinforcement learning for dependency parsing. They compare their results to \newcite{niv:hal:nil:ery:mar:06} using the same features, but they also change the model and apply beam search. It is thus unclear to what extend their improvements are due to reinforcement learning.

Even though most approaches mentioned above improve the results reported by \newcite{chen.manning.2014} and even more impressive results on dependency parsing have been achieved since (notably, \newcite{Andor2016}), Chen and Manning's parser provides a better baseline for our purposes. We aim at investigating the influence of reinforcement learning on error propagation and want to test this in a clean environment, where reinforcement learning does not interfere with other methods that address the same problem.

\subsection{Reinforcement Learning}

Reinforcement learning has been applied to several NLP tasks with  success, e.g.\ agenda-based parsing \cite{Jiang2012}, semantic parsing \cite{berant2015imitation} and simultaneous machine translation \cite{GrissomII2014}. To our knowledge, however, none of these studies investigated the influence of reinforcement learning on error propagation.

Learning to Search (L2S) is probably the most prominent line of research that applies reinforcement learning (more precisely, imitation learning) to NLP. Various algorithms, e.g.\ SEARN \cite{Daume09} and DAgger \cite{Ross2010}, have been developed sharing common high-level steps: a \textit{roll-in} policy is executed to generate training states from which a \textit{roll-out} policy is used to estimate the loss of certain actions. The concrete instantiation differs from one algorithm to another with choices including a referent policy (static or dynamic oracle), learned policy, or a mixture of the two. Early work in L2S focused on reducing reinforcement learning into binary classification \cite{Daume09}, but newer systems favored regressors for efficiency \cite[Supplementary material, Section B]{chang15}.
Our algorithm APG is simpler than L2S in that it uses only one policy (pre-trained with standard supervised learning) and applies the existing classifier directly without reduction (the only requirement is that it is probabilistic). Nevertheless, our results demonstrate its effectiveness. 

APG belongs to the family of policy gradient algorithms \cite{Sutton1999}, i.e.\ it maximizes the expected reward directly by following its gradient w.r.t.\ the parameters. 
The advantage of using a policy gradient algorithm in NLP is that gradient-based optimization is already widely used. REINFORCE \cite{williams92,Ranzato2016} is a widely-used policy gradient algorithm but it is also well-known for suffering from high variance \cite{Sutton1999}.

We directly compare our approach to REINFORCE, whereas we leave a direct comparison to L2S for future work. Our experiments show that our algorithm results in lower variance and achieves better performance than REINFORCE.

Recent work addresses the approximation of reinforcement learning gradient in the context of machine translation. \newcite{Shen2016}'s algorithm is roughly equivalent to the combination of an oracle and random sampling. Their approach differs from ours, because it does not retain memory across iteration as in our best-performing model (see Section~\ref{ssec:apg}).

\subsection{Reinforcement and error propagation}

As mentioned above, previous work that applied reinforcement learning to NLP has, to our knowledge, not shown that it improved results by reducing error propagation.

Work on identifying the impact of error propagation in parsing is rare, \newcite{Ng2015IdentifyingParsing} being a notable exception. They provide a detailed error analysis for parsing and classify which kind of parsing errors are involved with error propagation. There are four main differences between their approaches and ours. First, Ng and Curran correct arcs in the tree and our algorithm corrects decisions of the parsing algorithm. Second, our approach distinguishes between cases where one erroneous action deterministically leads to multiple erroneous arcs and cases where an erroneous action leads to conditions that indirectly result in further errors (see Section~\ref{ssec:errorprop} for a detailed explanation). Third, Ng and Curran's algorithm corrects all erroneous arcs that are the same type of parsing error and point out that they cannot examine the interaction between multiple errors of the same type in a sentence. Our algorithm corrects errors incrementally and therefore avoids this issue.
Finally, the classification and analysis presented in \newcite{Ng2015IdentifyingParsing} are more extensive and detailed than ours. Our algorithm can, however, easily be extended to perform similar analysis. 
Overall, Ng and Curran's approach for error analysis and ours are complementary. Combining them and applying them to various systems would form an interesting direction for future work.

\begin{table*}[ht]
\centering\small{
\begin{tabular}{|l|l|l|r|l|}
\hline
\textbf{Step} & \textbf{Transition} & \textbf{Stack} & \multicolumn{1}{l|}{\textbf{Buffer}} & \textbf{Arcs}\\
\hline
0 & & \sc{$<$root$>$} & \textit{waves hit ... Big Board} & $\emptyset$ \\
1 & {\sc shift} & \sc{$<$root$>$} \textit{waves} & \textit{hit stocks ... Big Board} & $\emptyset$ \\
2 & {\sc shift} & \sc{$<$root$>$} \textit{waves hit} & \textit{stocks themselves ... Big Board} & $\emptyset$ \\
3 & {\sc left}$\textsubscript{nsubj}$ & \sc{$<$root$>$} \textit{hit} & \textit{stocks themselves ... Big Board} & A$_1$ = \{ \textit{hit}
$\xrightarrow{\mathrm{nsubj}}$
\textit{waves}\} \\
4 & {\sc shift} &\sc{$<$root$>$} \textit{hit stocks} & \textit{themselves on the Big Board} & A$_1$ \\
5 & {\sc shift} & \sc{$<$root$>$} \textit{hit stocks themselves} & \textit{on the Big Board} & A$_1$ \\
6 & {\sc right}$\textsubscript{dep}$ & \sc{$<$root$>$} \textit{hit stocks} & \textit{on the Big Board} & A$_2$ = A$_1 \cup$  \\
 & & & & \{ \textit{stock} $\xrightarrow{\mathrm{dep}}$ \textit{themselves}\} \\
7 &  {\sc right}$\textsubscript{dobj}$ & \sc{$<$root$>$} \textit{hit} & \textit{on the Big Board} & A$_3$ = A$_2 \cup$ \{ \textit{hit} $\xrightarrow{\mathrm{dobj}}$ \textit{stock}\}  \\
\hline
\end{tabular}}
\caption{Parsing oracle walk-through}\label{tab:goldprocess}
\end{table*}

\section{A Reinforced Greedy Parser}\label{sec:ourparser}

This section describes the systems used in our experiments. We first describe the arc-standard algorithm, because familiarity with it helps to understand our error propagation analysis. Next, we briefly point out the main differences between the arc-standard algorithm and the alternative algorithms we experimented with (arc-eager and swap-standard). We then outline the traditional and our novel machine learning approaches. 
The features we used are identical to those described in \newcite{chen.manning.2014}. We are not aware of research identifying the best feature for a neural parser with arc-eager or swap-standard so we use the same features for all transition systems. 

\subsection{Transition-Based Dependency Parsing}
\label{ssec:parsing}

In an arc-standard system \cite{W04-0308}, a parsing configuration consists of a triple $\langle \Sigma, \beta, A \rangle$, where $\Sigma$ is a stack, $\beta$ is a buffer containing the remaining input tokens and $A$ are the dependency arcs that are created during parsing process. At initiation, the stack contains only the root symbol ($\Sigma$ = [{\tt ROOT}]), the buffer contains the tokens of the sentence ($\beta = [w_1,...,w_n]$) and the set of arcs is empty ($A = \emptyset$).

The arc-standard system supports three transitions. When $\sigma_1$ is the top element and $\sigma_2$ the second element on the stack, and $\beta_1$ the first element of the buffer:\footnote{Naturally, the transitions LEFT$_l$ and RIGHT$_l$ can only take place if the stack contains at least two elements and SHIFT can only occur when there is at least one element on the buffer.}

\begin{description}
\setlength{\itemsep}{-0.1cm}
\item[LEFT$_l$] adds an arc $\sigma_1 \xrightarrow{l} \sigma_2$ to $A$ and removes $\sigma_2$ from the stack.
\item[RIGHT$_l$] adds an arc $\sigma_2 \xrightarrow{l} \sigma_1$ to $A$ and removes $\sigma_1$ from the stack.
\item[SHIFT] moves $\beta_1$ to the stack.
\end{description}

\begin{figure}[t]
\centering
\includegraphics[scale=0.47]{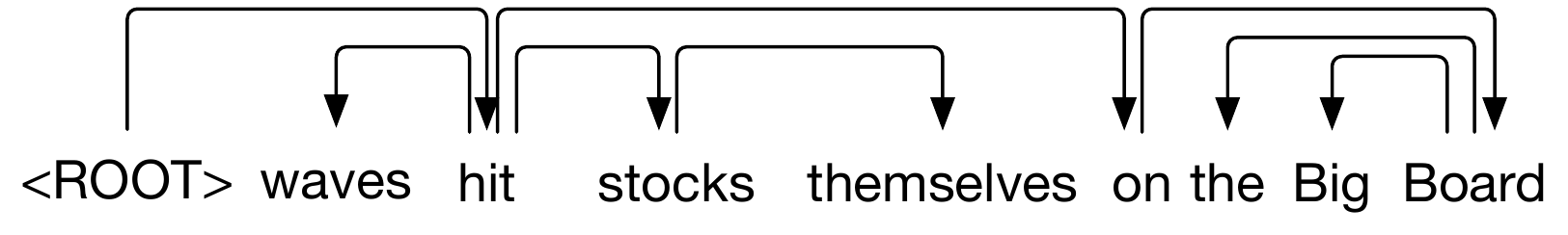}
\caption{Correct dependencies for a simplified example from Penn TreeBank}
\label{fig:correctdeps}
\end{figure}

When the buffer is empty, the stack contains only the root symbol and $A$ contains a parse tree, the configuration is completed. For a sentence of $N_w$ tokens, a full parse takes 2$N_w$ + 1 transitions to complete (including the initiation). Figure~\ref{fig:correctdeps} provides the gold parse tree for a (simplified) example from the Penn Treebank. The steps taken to create the dependencies between the sentence's head word \textit{hit} and its subject and direct object are provided in Table~\ref{tab:goldprocess}.

To demonstrate that reinforcement learning can train different systems, we also carried out experiments with arc-eager \cite{Nivre03anefficient} and swap-standard \cite{Nivre09}. Arc-eager is designed for incremental parsing and included in the popular MaltParser \cite{Nivre2006a}. Swap-standard is a simple and effective solution to unprojective dependency trees. Because arc-eager does not guarantee complete parse trees, we used a variation that employs an action called UNSHIFT to resume processing of tokens that would otherwise not be attached to a head \cite{nivre.gonzalez14}.

\subsection{Training with a Static Oracle}\label{ssec:statoracle}

In transition-based dependency parsing, it is common to convert a dependency treebank $\mathcal{D} \ni (x,y)$ into a collection of input features $s \in \mathcal{S}$ and corresponding gold-standard actions $a \in \mathcal{A}$ for training, using a static oracle $\mathcal{O}$. In \newcite{chen.manning.2014}, a neural network works as a function mapping input features to probabilities of actions: $f_{NN}: \mathcal{S} \times \mathcal{A} \rightarrow [0,1]$. The neural network is trained to minimize negative log-likelihood loss on the converted treebank:
\begin{equation}
\mathcal{L} = \sum_{(x,y) \in \mathcal{D}} \sum_{(s, a) \in \mathcal{O}(x,y)} - \log f_{NN}(s,a; \theta)
\end{equation}

\subsection{Reinforcement Learning}
\label{sec:rl}

Following \newcite{Maes2009}, we view transition-based dependency parsing as a deterministic Markov Decision Process. The problem is summarized by a tuple $\langle \mathcal{S}, \mathcal{A}, \mathcal{T}, r \rangle$ where $\mathcal{S}$ is the set of all possible states, $\mathcal{A}$ contains all possible actions, $\mathcal{T}$ is a mapping $\mathcal{S} \times \mathcal{A} \rightarrow \mathcal{S}$ called \textit{transition function} and $r: \mathcal{S} \times \mathcal{A} \rightarrow \mathbb{R}$ is a reward function.

A state corresponds to a configuration and is summarized into input features. Possible actions are defined for each transition system described in Section~\ref{ssec:parsing}. We keep the training approach simple by using only one reward $r(\bar{y})$ at the end of each parse.

Given this framework, a stochastic policy guides our parser by mapping each state to a probabilistic distribution of actions. During training, we use function $f_{NN}$ described in Section~\ref{ssec:statoracle} as a stochastic policy. At test time, actions are chosen in a greedy fashion following existing literature. We aim at finding the policy that maximizes the expected reward (or, equivalently, minimizes the expected loss) on the training dataset:
\begin{equation}
\text{maximize}\ \eta = \sum_{(x,y) \in \mathcal{D}} \sum_{a_{1:m} \sim f} r(\bar{y}) \prod_{i=1}^m f_{NN}(s_i, a_i; \theta)
\label{eq:maximize-eta}
\end{equation}
where $a_{1:m}$ is a sequence of actions obtained by following policy $f_{NN}$ until termination and $s_{1:m}$ are corresponding states (with $s_{m+1}$ being the termination state).

\subsection{Approximate Policy Gradient}
\label{ssec:apg}

Gradient ascent can be used to maximize the expected reward in Equation~\ref{eq:maximize-eta}.
The gradient of expected reward w.r.t.\ parameters is (note that $\mathrm{d} z = z \mathrm{d}(\log z)$):
\begin{equation}
\begin{split}
\frac{\partial \eta}{\partial \theta} = & \sum_{(x,y) \in \mathcal{D}} \sum_{a_{1:m} \sim f_{NN}} r(\bar{y}) \prod_{i=1}^m f_{NN}(s_i, a_i) \\ 
 & \sum_{i=1}^m \frac{\partial}{\partial \theta} \log f_{NN}(s_i, a_i; \theta)
\label{eq:true-gradient}
\end{split}
\end{equation}

Because of the exponential number of possible trajectories, calculating the gradient exactly is not possible. We propose to replace it by an approximation (hence the name \textit{Approximate} Policy Gradient) by summing over a small subset $U$ of trajectories. Following common practice, we also use a baseline $b(y)$ that only depends on the correct dependency tree. The parameter is then updated by following the approximate gradient:
\begin{equation}
\begin{split}
\Delta \theta \propto & \sum_{(x,y) \in \mathcal{D}} \sum_{a_{1:m} \in U} (r(\bar{y})-b(y)) \prod_{i=1}^m f_{NN}(s_i, a_i) \\ 
 & \sum_{i=1}^m \frac{\partial}{\partial \theta} \log f_{NN}(s_i, a_i; \theta)
\label{eq:gradient-rl}
\end{split}
\end{equation}

Instead of sampling one trajectory at a time as in REINFORCE, Equation~\ref{eq:gradient-rl} has the advantage that sampling over multiple trajectories could lead to more stable training and higher performance. To achieve that goal, the choice of $U$ is critical. We empirically evaluate three strategies:
\begin{description}
\setlength{\itemsep}{-0.1cm}
\item[\textsc{RL-Oracle}:] only includes the oracle transition sequence.
\item[\textsc{RL-Random}:] randomly samples $k$ distinct trajectories at each iteration. Every action is sampled according to $f_{NN}$, i.e.\ preferring trajectories for which the current policy assigns higher probability.
\item[\textsc{RL-Memory}:] samples randomly as the previous method but retains $k$ trajectories with highest rewards across iterations in a separate memory. Trajectories are ``forgotten'' (removed) randomly with probability $\rho$ before each iteration.\footnote{We assign a random number (drawn uniformly from $[0, 1]$) to each trajectory in memory and remove those assigned a number less than $\rho$.}
\end{description}

Intuitively, trajectories that are more likely and produce higher rewards are better training examples. It follows from Equation~\ref{eq:true-gradient} that they also bear bigger weight on the true gradient. This is the rationale behind \textsc{RL-Random} and \textsc{RL-Oracle}. For a suboptimal parser, however, these objectives sometimes work against each other. \textsc{RL-Memory} was designed to find the right balance between them. It is furthermore important that the parser is pretrained to ensure good samples.
Algorithm~\ref{alg:apg} illustrates the procedure of training a dependency parser using the proposed algorithms.

\begin{algorithm}[t]
\SetAlgoLined
 $MemorySeqs \leftarrow \emptyset$\;
 \ForEach{training batch b}{
 \ForEach{sentence s $\in$ b}{
  $OracleSeq \leftarrow$ Oracle($s$)\;
  $SystemSeqs \leftarrow$ (sample $k$ parsing transition sequences for $s$)\;
  \uIf{RL-Oracle}{
    ComputeGradients($OracleSeq$)\;
  }\uElseIf{RL-Random}{
    ComputeGradients($SystemSeqs$)\;
  }\uElseIf{RL-Memory}{
  	$m \leftarrow MemorySeqs[s]$\;
    \ForEach{q $\in$ m}{
    	\If{RandomNumber() < $\rho$}{
        	Remove $q$ from $m$\;
        }
    }
    \ForEach{q $\in$ SystemSeqs}{
    	\uIf{|m| < k}{
        	Insert $q$ into $m$\;
        } \uElse {
          $p \leftarrow $ (sequence with smallest reward in $m$)\;
          \If{reward(q) > reward(p)}{
              Replace $p$ by $q$ in $m$\;
          }        
        }
    }
    ComputeGradients($m$)\;
  }
  }
  Perform one gradient descent step\;
 }
 \caption{Training a dependency parser with approximate policy gradient.}
 \label{alg:apg}
\end{algorithm}

\section{Reinforcement Learning Experiments}
\label{sec:reinforcement-experiments}


We first train a parser using a supervised learning procedure and then improve its performance using APG. We empirically tested that training a second time with supervised learning has little to no effect on performance.

\subsection{Experimental Setup}\label{ssec:experimental-setup}

We use PENN Treebank 3 with standard split (training, development and test set) for our experiments with arg-standard and arg-eager. Because the swap-standard parser is mainly suited for non-projective structures, which are rare in the PENN Treebank, we evaluate this parser on the German section of the SPMRL dataset.
For PENN Treebank, we follow Chen and Manning's preprocessing steps. We also use their pretrained model\footnote{We use \texttt{PTB\_Stanford\_params.txt.gz} downloaded from \texttt{http://nlp.stanford.edu/software/\allowbreak nndep.\allowbreak shtml} on December 30\textsuperscript{th}, 2015.} for arc-standard and train our own models in similar settings for other transition systems.

For reinforcement learning
, we use AdaGrad for optimization. We do not use dropout because we observed that it destablized the training process. The reward $r(\bar{y})$ is the number of correct labeled arcs (i.e.\ LAS multiplied by number of tokens).\footnote{Punctuation is not taken into account, following \newcite{chen.manning.2014}.} The baseline is fixed to half the number of tokens (corresponding to a 0.5 LAS score). As training takes a lot of time, we tried only few values of hyperparameters on the development set and picked $k = 8$ and $\rho = 0.01$. 1,000 updates were performed (except for \textsc{REINFORCE} which was trained for 8,000 updates) with each training batch contains 512 randomly selected sentences. The Stanford dependency scorer\footnote{Downloaded from \texttt{http://nlp.stanford.edu/\allowbreak software/\allowbreak lex-parser.shtml}.} was used for evaluation.
%

\subsection{Effectiveness of Reinforcement Learning}
\label{ssec:rl-effectiveness}

\begin{table}
\centering\small{
\begin{tabular}{|l|c|c|c|c|c|c|}
\hline
& \multicolumn{2}{c|}{\textbf{Arc-}} & \multicolumn{2}{c|}{\textbf{Arc-}} & \multicolumn{2}{c|}{\textbf{Swap-}} \\ 
& \multicolumn{2}{c|}{\textbf{standard}} & \multicolumn{2}{c|}{\textbf{eager}} & \multicolumn{2}{c|}{\textbf{standard}} \\ \cline{2-7}
& \textbf{UAS} & \textbf{LAS} & \textbf{UAS} & \textbf{LAS} & \textbf{UAS} & \textbf{LAS} \\ \hline
\textsc{SL} & 91.3 & 89.4 & 88.3 & 85.8 & 84.3 & 81.3 \\ \hline
\textsc{RE} & 91.9 & 90.2 & 89.7 & 87.2 & 87.5 & 84.4 \\ \hline
\textsc{RL-O} & 91.8 & 90.2 & 88.9 & 86.5 & 86.8 & 83.9 \\ \hline
\textsc{RL-R} & \textbf{92.2} & \textbf{90.6} & 89.4 & 87.0 & 87.5 & 84.5 \\ \hline
\textsc{RL-M} & \textbf{92.2} & \textbf{90.6} & \textbf{89.8} & \textbf{87.4} & \textbf{87.6} & \textbf{84.6} \\ \hline
\end{tabular}}
\caption{Comparing training methods on PENN Treebank (arc-standard and arc-eager) and German part of SPMRL-2014 (swap-standard).}
\label{tab:accuracy}
\end{table}

Table~\ref{tab:accuracy} displays the performance of different approaches to training dependency parsers. Although we used \newcite{chen.manning.2014}'s pretrained model and Stanford open-source software, the results of our baseline are slightly worse than what is reported in their paper. This could be due to minor differences in settings and does not affect our conclusions. 

Across transition systems and two languages, APG outperforms supervised learning, verifying our hypothesis. Moreover, it is not simply because the learners are exposed to more examples than their supervised counterparts. \textsc{RL-Oracle} is trained on exactly the same examples as the standard supervised learning system (\textsc{SL}), yet it is consistently superior. This can only be explained by the superiority of the reinforcement learning objective function compared to negative log-likelihood.

The results support our hypothesis that \textsc{APG} is better than \textsc{REINFORCE} (abbreviated as \textsc{RE} in Table~\ref{tab:accuracy}) as \textsc{RL-Memory} always outperforms the classical algorithm and the other two heuristics do in two out of three cases. The usefulness of training examples that contain errors is evident through the better performance of \textsc{RL-Random} and \textsc{RL-Memory} in comparison to \textsc{RL-Oracle}.

Table~\ref{tab:sample-size} shows the importance of samples for \textsc{RL-Random}. The algorithm hurts performance when only one sample is used whereas training with two or more samples improves the results. The difference cannot be explained by the total number of observed samples because one-sample training is still worse after 8,000 iterations compared to a sample size of 8 after 1,000 iterations.
The benefit of added samples is twofold: increased performance and decreased variance. Because these benefits saturate quickly, we did not test sample sizes beyond 32.

\newcommand{\s}{$^*$}

\begin{table}[h]\centering\small{
\begin{tabular}{|c|c|c|c|c||c|c|}
\hline
& \multicolumn{2}{|c|}{\textbf{Dev}} & \multicolumn{2}{|c||}{\textbf{Test}} & \multicolumn{2}{|c|}{\textbf{Test std.}}\\ \cline{2-7}
& \textbf{UAS} & \textbf{LAS} & \textbf{UAS} & \textbf{LAS} & \textbf{UAS} & \textbf{LAS} \\ \hline
SL & 91.5 & 89.6 & 91.3 & 89.4 & - & - \\ \hline
RE & 92.1\s & 90.4\s & 91.9\s & 90.2\s & 0.04 & 0.05 \\ \hline
1 & 91.2\s & 89.1\s & 91.0\s & 88.9\s & 0.12 & 0.15 \\ \hline
2 & 91.8\s & 90.0\s & 91.6\s & 89.9\s & 0.09 & 0.09 \\ \hline
4 & 92.2\s & 90.5\s & 92.0\s & 90.4\s & 0.09 & 0.08 \\ \hline
8 & \textbf{92.4}\s & \textbf{90.8}\s & 92.2\s & \textbf{90.6}\s & 0.03 & 0.05 \\ \hline
16 & \textbf{92.4} & \textbf{90.8} & 92.2 & \textbf{90.6} & - & - \\ \hline
32 & \textbf{92.4} & \textbf{90.8} & \textbf{92.3} & \textbf{90.6} & - & - \\ \hline
\end{tabular}
\caption{Parsing accuracy of \textsc{RL-Random} (arc-standard) with different sample sizes compared to supervised learning (SL) and REINFORCE (RE). \s: significantly different from \textsc{SL} with $p < 0.001$}
\label{tab:sample-size}}
\end{table}

\section{Error Propagation Experiment}\label{sec:errorpropexp}

We hypothesized that reinforcement learning avoids error propagation. In this section, we describe our algorithm and the experiment that identifies error propagation in the arc-standard parsers.

\subsection{Error Propagation}\label{ssec:errorprop}

\begin{table*}
\centering\small{
\begin{tabular}{|l|l|l|r|l|}
\hline
\textbf{Step} & \textbf{Transition} & \textbf{Stack} & \multicolumn{1}{l|}{\textbf{Buffer}} & \textbf{Arcs}\\
\hline
4 & {\sc shift} & {\sc $<$root$>$} \textit{hit stocks} & \textit{themselves on the Big Board} & A$_1$ \\
5' & {\sc right}\textsubscript{dobj} &  {\sc $<$root$>$} \textit{hit} & \textit{themselves on the Big Board} & A$_2$ = A$_1 \cup$ \\
& & & & \{\textit{hit} $\xrightarrow{\mathrm{dobj}}$ \textit{stock}\} \\
6' & {\sc shift} & {\sc $<$root$>$} \textit{hit themselves} & \textit{on the Big Board} & A$_2$  \\
7' & {\sc shift} &  {\sc $<$root$>$} \textit{hit themselves on} & \textit{the Big Board} & A$_2$\\
... & & & & \\
10' & {\sc shift} &  {\sc $<$root$>$} \textit{hit themselves on the Big Board} & & A$_2$ \\
11' & {\sc left}\textsubscript{nn} & {\sc $<$root$>$} \textit{hit themselves on the Board} & & A$_3$ = A$_2 \cup$ \\
& & & & \{\textit{Board} $\xrightarrow{\mathrm{nn}}$ \textit{Big}\}\\
12' & {\sc left}\textsubscript{det} &  {\sc $<$root$>$} \textit{hit themselves on Board} & & A$_4$ = A$_3 \cup$ \\
& & & & \{\textit{Board} $\xrightarrow{\mathrm{det}}$ \textit{the}\} \\
13' & {\sc right}\textsubscript{pobj} &  {\sc $<$root$>$} \textit{hit themselves on} & & A$_5$ = A$_4 \cup$ \\
& & & & \{\textit{on} $\xrightarrow{\mathrm{pobj}}$ \textit{Board}\} \\
14' & {\sc right}\textsubscript{dep} & {\sc $<$root$>$} \textit{hit themselves} & & A$_6$ = A$_5 \cup$ \\
& & & & \{\textit{themselves} $\xrightarrow{\mathrm{dep}}$\textit{on}\}\\
\hline
\end{tabular}}
\caption{Possible parsing walk-through with error}\label{tab:actualprocess}
\end{table*}

Section~\ref{ssec:parsing} explained that a transition-based parser goes through the sentence incrementally and must select a transition from [SHIFT, LEFT$_l$, RIGHT$_l$] at each step. We use the term \textit{arc error} to refer to an erroneous arc in the resulting tree. The term \textit{decision error} refers to a transition that leads to a loss in parsing accuracy. Decision errors in the parsing process lead to one or more arc errors in the resulting tree. There are two ways in which a single decision error may lead to multiple arc errors. First, the decision can deterministically 
lead to more than one arc error, because (e.g.) an erroneously formed arc also blocks other correct arcs.
Second, an erroneous parse decision changes some of the features that the model uses for future decisions and these changes can lead to further (decision) errors down the road.



\begin{figure}
\centering
\includegraphics[scale=0.43]{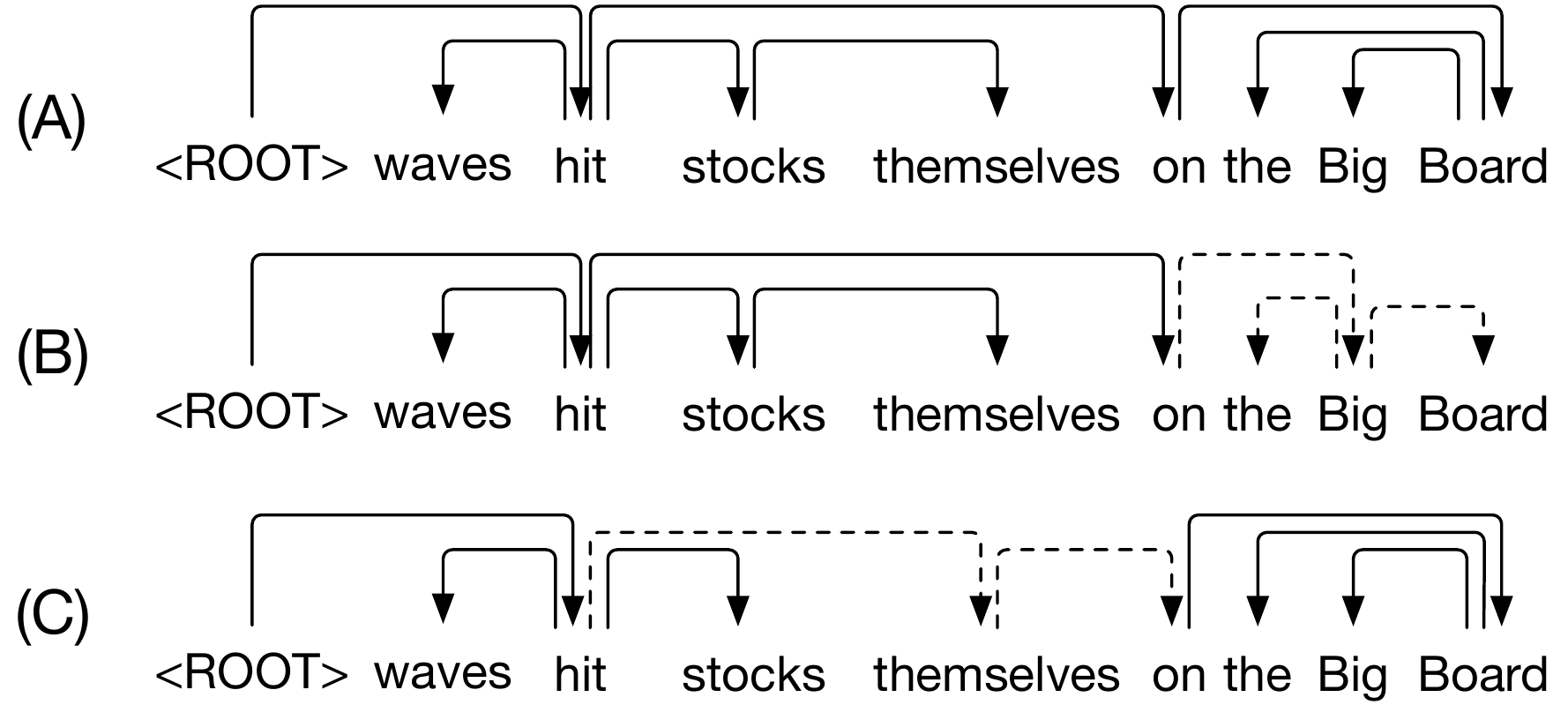}
\caption{Three dependency graphs: gold (A), arc errors caused by one decision error (B) and arc errors caused by multiple decision errors (C).}
\label{fig:alldeps}
\end{figure}

We illustrate both cases using two incorrect derivations presented in Figure~\ref{fig:alldeps}. The original gold tree is repeated in (A). The dependency graph in Figure~\ref{fig:alldeps} (B) contains three erroneous dependency arcs (indicated by dashed arrows). The first error must have occurred when the parser executed RIGHT\textsubscript{amod} creating the arc \textit{Big} $\rightarrow$ \textit{Board}. After this error, it is impossible to create the correct relations \textit{on} $\rightarrow$ \textit{Board} and \textit{Board} $\rightarrow$ \textit{the}. The wrong arcs \textit{Big} $\rightarrow$ \textit{the} and \textit{on} $\rightarrow$ \textit{Big} are thus all the result of a single decision error. 

Figure~\ref{fig:alldeps} (C) represents the dependency graph that is actually produced by our parser.\footnote{The example is a fragment of a more complex sentence consisting of 33 tokens. The parser does provide the correct output when is analyzes this sequence in isolation.} It contains two erroneous arcs: \textit{hit} $\rightarrow$ \textit{themselves} and \textit{themselves} $\rightarrow$ \textit{on}. Table~\ref{tab:actualprocess} provides a possible sequence of steps that led to this derivation, starting from the moment \textit{stocks} was added to the stack (Step 4). The first error is introduced in Step 5', where \textit{hit} combines with \textit{stocks} before \textit{stocks} has picked up its dependent \textit{themselves}. At that point, \textit{themselves} can no longer be combined with the right head. The proposition \textit{on}, on the other hand, can still be combined with the correct head. This error is introduced in Step 7', where the parser moves \textit{on} to the stack rather than creating an arc from \textit{hit} to \textit{themselves}.\footnote{Note that technically, \textit{on} can still become a dependent of \textit{hit}, but this can only happen if \textit{on} becomes the head of \textit{themselves} which would also be an error.} There are thus two decision errors that lead to the arc errors in Figure~\ref{fig:alldeps} (C). The second decision error can, however, be caused indirectly by the first error. If a decision error causes additional decision errors later in the parsing process, we talk of error propagation. This cannot be known just by looking at the derivation. 

\subsection{Examining the impact of decision errors}\label{ssec:errorpropid}

We examine the impact of individual decision errors on the overall parse results in our test set by combining a dynamic oracle and a recursive function. We use a dynamic oracle based on \newcite{goldberg2014} 
which gives us the overall loss at any point during the derivation. The loss is equal to 
the minimal number of arc errors that will have been made once the parse is complete. 
We can thus deduce how many arc errors are deterministically caused by a given decision error. 


The propagation of decision errors cannot be determined by simply examining the increase in loss during the parsing process. We use a recursive function to identify whether a particular parse suffered from this. While parsing the sentence, we register which decisions lead to an increase in loss. We then recursively reparse the sentence correcting one additional decision error during each run until the parser produces the gold. If each erroneous decision has to be corrected in order to arrive at the gold, we assume the decision errors are independent of each other. If, on the other hand, the correction of a specific decision also fixes other decisions down the road, the original parse suffers from error propagation.

\begin{table}
\centering{\fontsize{9.4}{11.4}
\begin{tabular}{|l|c|c|c|c|}
\hline
& \textbf{\textsc{SL}} & \textbf{\textsc{RL-O}} & \textbf{\textsc{RL-R}} & \textbf{\textsc{RL-M}} \\
\hline
Total Loss & 7069 & 6227 & 6042 & 6144 \\
Dec.\ Errors & 5177 & 4410 & 4345 & 4476 \\
Err.\ Prop. & 1399 & 1124 & 992 & 1035 \\
New errors & 411 & 432 & 403 & 400 \\
\hline
Loss/error & 1.37 & 1.41 & 1.39 & 1.37 \\
Err.\ Prop. (\%) & 27.0 & 25.5 & 22.8 & 23.1 \\
\hline
\end{tabular}
\caption{Overview of average impact of decision errors}\label{tab:errorprop}
}
\end{table}


The results are presented in Table~\ref{tab:errorprop}. \textit{Total Loss} indicates the number of arc errors in the corpus, \textit{Dec.\ Errors} the number of decision errors and \textit{Err.\ Prop.} the number of decision errors that were the result of error propagation. This number was obtained by comparing the number of decision errors in the original parse to the number of decision errors that needed to be fixed to obtain the gold parse. If less errors had to be fixed than originally present, we counted the difference as error propagation. Note that fixing errors sometimes leads to new decision errors during the derivation. We also counted the cases where more decision errors needed to be fixed than were originally present and report them in Table~\ref{tab:errorprop}.\footnote{We ran an alternative analysis where we counted all cases where fixing one decision error in the derivation reduced the overall number of decision errors in the parse by more than one. Under this alternative analysis, similar reductions in the proportion of error propagation were observed for reinforcement learning.}

On average, decision errors deterministically lead to more than one arc error in the resulting parse tree. This remains stable across systems (around 1.4 arc errors per decision error). We furthermore observe that the proportion of decision errors that are the result of error propagation has indeed reduced for all reinforcement learning models.
Among the errors avoided by APG, 35.9\% were propagated errors for \textsc{RL-Oracle}, 48.9\% for \textsc{RL-Random}, and 51.9\% for \textsc{RL-Memory}. These percentages are all higher than the proportion of propagated errors occurring in the corpus parsed by \textsc{SL} (27\%).
This outcome confirms our hypothesis that reinforcement learning is indeed more robust for making decisions in imperfect environments and therefore reduces error propagation. 

%
%

\section{Conclusion}\label{sec:conclusion}

This paper introduced Approximate Policy Gradient (APG), an efficient reinforcement learning algorithm for NLP, and applied it to a high-performance greedy dependency parser. We hypothesized that reinforcement learning would be more robust against error propagation and would hence improve parsing accuracy. 

To verify our hypothesis, we ran experiments applying APG to three transition systems and two languages. We furthermore introduced an experiment to investigate which portion of errors were the result of error propagation and compared the output of standard supervised machine learning to reinforcement learning.
Our results showed that: (a) reinforcement learning indeed improved parsing accuracy and (b) propagated errors were over-represented in the set of avoided errors, confirming our hypothesis. 

To our knowledge, this paper is the first to show experimentally that reinforcement learning can reduce error propagation in an NLP task. This result was obtained by a straight-forward implementation of reinforcement learning. Furthermore, we only applied reinforcement learning in the training phase, leaving the original efficiency of the model intact. Overall, we see the outcome of our experiments as an important first step in exploring the possibilities of reinforcement learning for tackling error propagation. 

Recent research on parsing has seen impressive improvement during the last year achieving UAS around 94\% \cite{Andor2016}. This improvement is partially due to other approaches that, at least in theory, address error propagation, such as beam search. Both the reinforcement learning algorithm and the error propagation study we developed can be applied to other parsing approaches. There are two (related) main questions to be addressed in future work in the domain of parsing. The first addresses whether our method is complementary to alternative approaches and could also improve the current state-of-the-art. The second question would address the impact of various approaches on error propagation and the kind of errors they manage to avoid (following \newcite{Ng2015IdentifyingParsing}). 

APG is general enough for other structured prediction problems. We therefore plan to investigate whether we can apply our approach to other NLP tasks such as coreference resolution or semantic role labeling and investigate if it can also reduce error propagation for these tasks. 



The source code of all experiments is publicly available at \url{https://bitbucket.org/cltl/redep-java}.

\section*{Acknowledgments}

The research for this paper was supported by the Netherlands Organisation for Scientific Research (NWO) via the Spinoza-prize Vossen projects (SPI 30-673, 2014-2019) and the VENI project \textit{Reading between the lines} (VENI 275-89-029).
Experiments were carried out on the Dutch national e-infrastructure with the support of SURF Cooperative. 
We would like to thank our friends and colleagues Piek Vossen, Roser Morante, Tommaso Caselli, Emiel van Miltenburg, and Ngoc Do for many useful comments and discussions.
We would like to extend our thanks the anonymous reviewers for their feedback which helped improving this paper.  All remaining errors are our own.

\bibliography{Mendeley-Minh,Mendeley-Antske}
\bibliographystyle{acl}

\end{document}